\documentclass[10pt,twocolumn,letterpaper]{article}

\usepackage{cvpr}
\usepackage{times}
\usepackage{epsfig}
\usepackage{graphicx}
\usepackage{amsmath}
\usepackage{amssymb}
\usepackage{booktabs} 
\usepackage{multirow} 
\usepackage{float}
\usepackage{algorithmic}
\usepackage{algorithm}
\usepackage{bbm}
\usepackage[breaklinks=true,bookmarks=false]{hyperref}

\cvprfinalcopy 

\def\cvprPaperID{****} 

\begin{document}

\title{ToLo: A Two-Stage, Training-Free Layout-To-Image Generation Framework For High-Overlap Layouts}

\author{Linhao Huang\\
Beijing University of Technology\\
{\tt\small huanglinhao@emails.bjut.edu.cn}
\and
Jing Yu*\\
Beijing University of Technology\\
{\tt\small jing.yu@bjut.edu.cn}
}

\twocolumn[{%
\maketitle
\begin{figure}[H]
\hsize=\textwidth 
\centering
\includegraphics[width=2\linewidth]{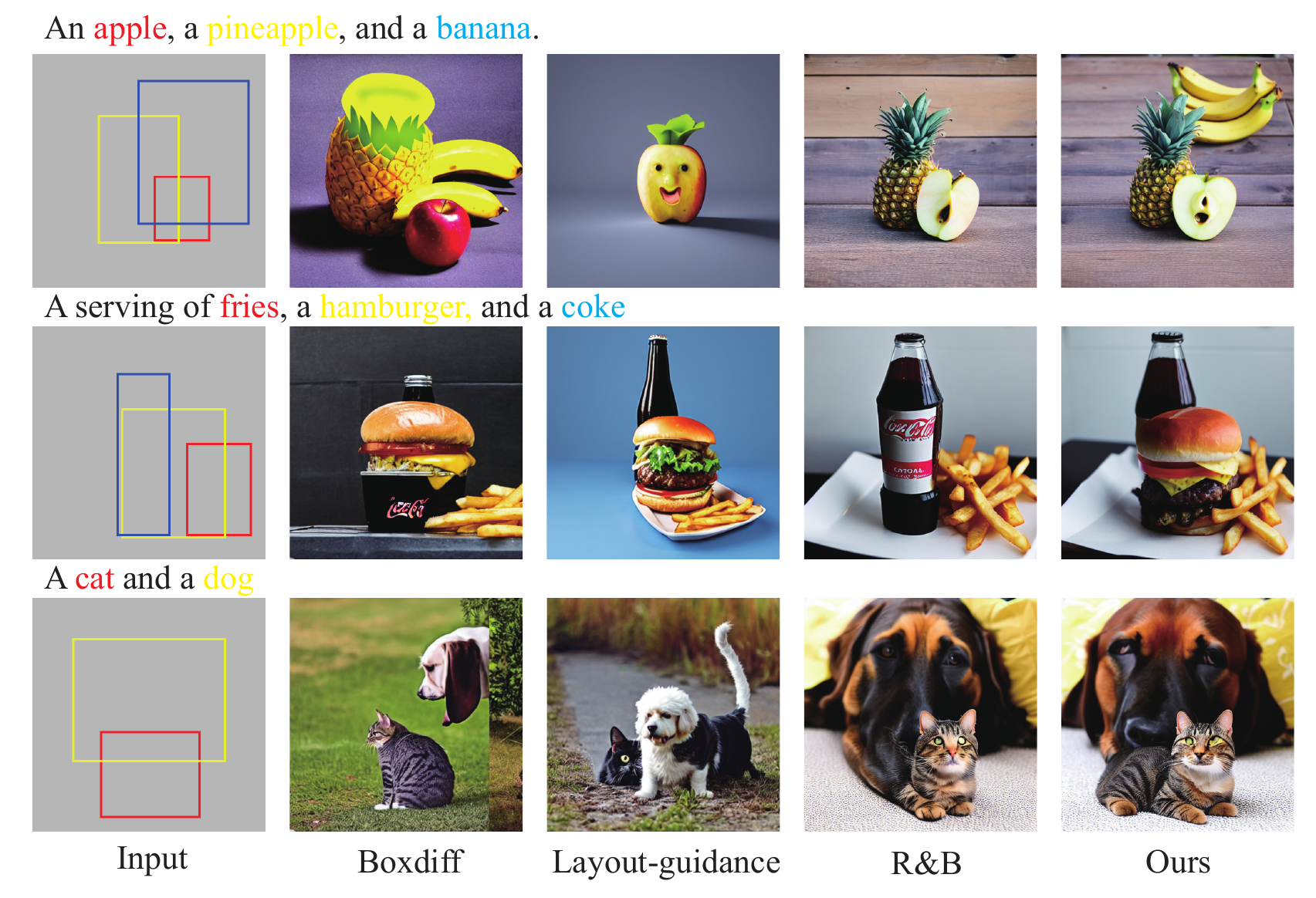}
\caption{Existing training-free layout-to-image synthesis methods struggle when the input layout contains large overlaps. Since they do not separate the attention maps for different objects, this often results in overlapping attention maps, which can cause the attribute leakage and missing entities. ToLo alleviates the problem of attribute leakage and missing entities while maintaining precise spatial control.}
\label{qualitative}
\end{figure}
}]

\begin{abstract}
Recent training-free layout-to-image diffusion models have demonstrated remarkable performance in generating high-quality images with controllable layouts. These models follow a one-stage framework: Encouraging the model to focus the attention map of each concept on its corresponding region by defining attention map-based losses. However, these models still struggle to accurately follow layouts with significant overlap, often leading to issues like attribute leakage and missing entities. In this paper, we propose ToLo, a two-stage, training-free layout-to-image generation framework for high-overlap layouts. Our framework consists of two stages: the aggregation stage and the separation stage, each with its own loss function based on the attention map. To provide a more effective evaluation, we partition the HRS dataset based on the Intersection over Union (IoU) of the input layouts, creating a new dataset for layout-to-image generation with varying levels of overlap. Through extensive experiments on this dataset, we demonstrate that ToLo significantly enhances the performance of existing methods when dealing with high-overlap layouts. Our code and dataset are available here: \href{https://github.com/misaka12435/ToLo}{https://github.com/misaka12435/ToLo}.
\end{abstract}

\section{Introduction}

In recent years, text-to-image synthesis models such as DALL-E \cite{Zero-ShotText-to-ImageGeneration}, Imagen \cite{PhotorealisticText-to-ImageDiffusionModelswithDeepLanguageUnderstanding}, and Stable Diffusion \cite{High-ResolutionImageSynthesisWithLatentDiffusionModels} have demonstrated an unprecedented ability to generate photorealistic, high-resolution images, thereby lowering the barriers to art and image creation.

Despite their success, text-based models often lack fine-grained control over the spatial structure of the generated images. Additionally, when an image contains multiple objects, text-only models struggle to assign the correct attributes to each object. These capabilities are essential for more sophisticated image generation applications. While text can convey a vast library of high-level concepts, it struggles to capture the subtle visual nuances inherent in images\cite{Training-FreeLayoutControlWithCross-AttentionGuidance}.

To address these challenges, several methods have been proposed that incorporate spatial layout as input for image generation, known as layout-to-image synthesis (LIS) \cite{Training-FreeLayoutControlWithCross-AttentionGuidance,GLIGEN:Open-SetGroundedText-to-ImageGeneration,GroundedText-to-ImageSynthesiswithAttentionRefocusing,LayeredRenderingDiffusionModelforControllableZero-ShotImageSynthesis,RnB:RegionandBoundaryAwareZero-shotGroundedText-to-imageGeneration,AddingConditionalControltoText-to-ImageDiffusionModels,LoCo:LocallyConstrainedTraining-FreeLayout-to-ImageSynthesis}. These methods have shown promising results. Some approaches impose spatial constraints by introducing new modules, while keeping the parameters of pre-trained models unchanged and training only the new modules. For instance, ControlNet \cite{AddingConditionalControltoText-to-ImageDiffusionModels} employs such spatial controls, but this method can be computationally expensive and requires additional layout-image training data.

\begin{figure}[t]
\begin{center}
   \includegraphics[width=1\linewidth]{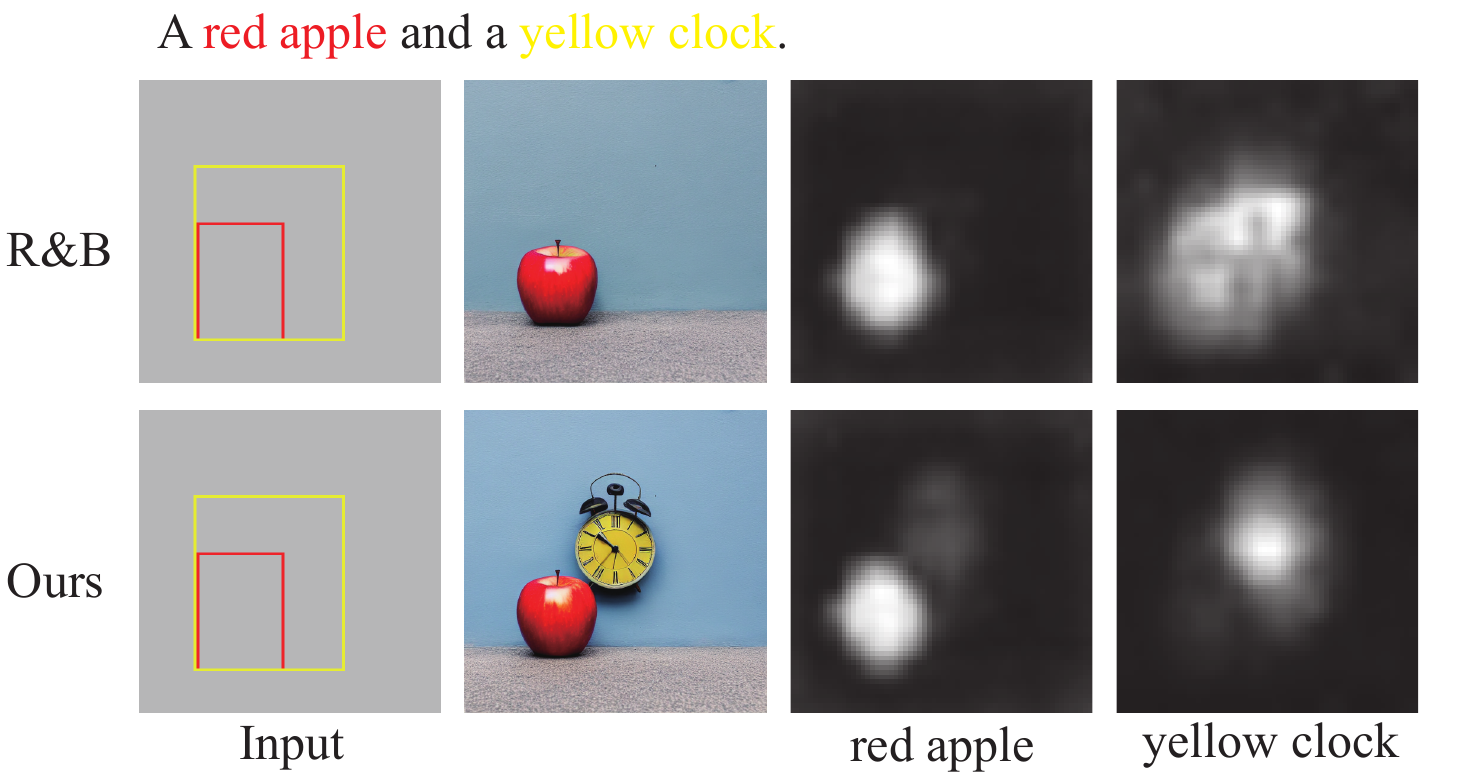}
\end{center}
   \caption{Attention maps of "red apple" and "yellow clock". In the original R\&B, the attention regions for the entities "red apple" and "yellow clock" overlap, with the attention map for the "yellow clock" being indistinguishable from that of the "red apple", resulting in the entity being missed (i.e., the "yellow clock" is omitted). However, after applying ToLo, the overlap is significantly alleviated, thereby the prompt "A red apple and a yellow clock" is correctly generated. More examples can be found in the Appendix \ref{appendix_attentionmap}.} 
\label{attentionmap}
\end{figure}

In contrast, other approaches\cite{Training-FreeLayoutControlWithCross-AttentionGuidance,GroundedText-to-ImageSynthesiswithAttentionRefocusing,RnB:RegionandBoundaryAwareZero-shotGroundedText-to-imageGeneration,LoCo:LocallyConstrainedTraining-FreeLayout-to-ImageSynthesis} tackle these problems in a training-free manner by manipulating the attention map. These methods encourage the model to focus the attention map of each concept on its corresponding region by defining attention map-based losses. They then use gradient descent to update the latent features during the inference phase, achieving layout control. Despite their success, these methods struggle when the input layout contains significant overlaps, often leading to issues such as attribute leakage and missing entities, as shown in Figure \ref{qualitative}. We observe that in these methods, overlapping layouts tend to result in overlapping attention maps, which may cause the aforementioned problems, as illustrated in Figure \ref{attentionmap}.

Based on this observation, we propose ToLo, a two-stage training-free layout-to-image generation framework, which consists of two stages: the aggregation stage and the separation stage. We define two layout-based loss functions, $L_{\rm agg}$ and $L_{\rm sep}$, for these stages. $L_{\rm agg}$ encourages the attention map of each concept to focus on its respective target region, achieving precise control over the spatial position of the concept. In contrast, $L_{\rm sep}$ ensures the separation between attention maps for different concepts, thereby mitigating the missing of entities and attribute leakage. We apply this new framework to R\&B \cite{RnB:RegionandBoundaryAwareZero-shotGroundedText-to-imageGeneration}, a state-of-the-art training-free layout-to-image generation method, and demonstrate the effectiveness and robustness of our framework through extensive experiments.

The main contributions of this paper are as follows:

\begin{itemize} \item We propose ToLo, a two-stage, training-free layout-to-image generation framework, comprising the aggregation and separation stages, each with a loss function. \item We apply our framework to R\&B \cite{RnB:RegionandBoundaryAwareZero-shotGroundedText-to-imageGeneration}, improving its performance on layouts with high overlap by adding only a few lines of code.  \item We partition the HRS dataset according to the IoU of the input layout, creating a new dataset for layout-to-image generation with varying levels of overlap. The effectiveness and robustness of our framework is demonstrated through comprehensive experiments on this
dataset. \end{itemize}

\section{Related Work}

\begin{figure*}
\begin{center}
\includegraphics[width=1\linewidth]{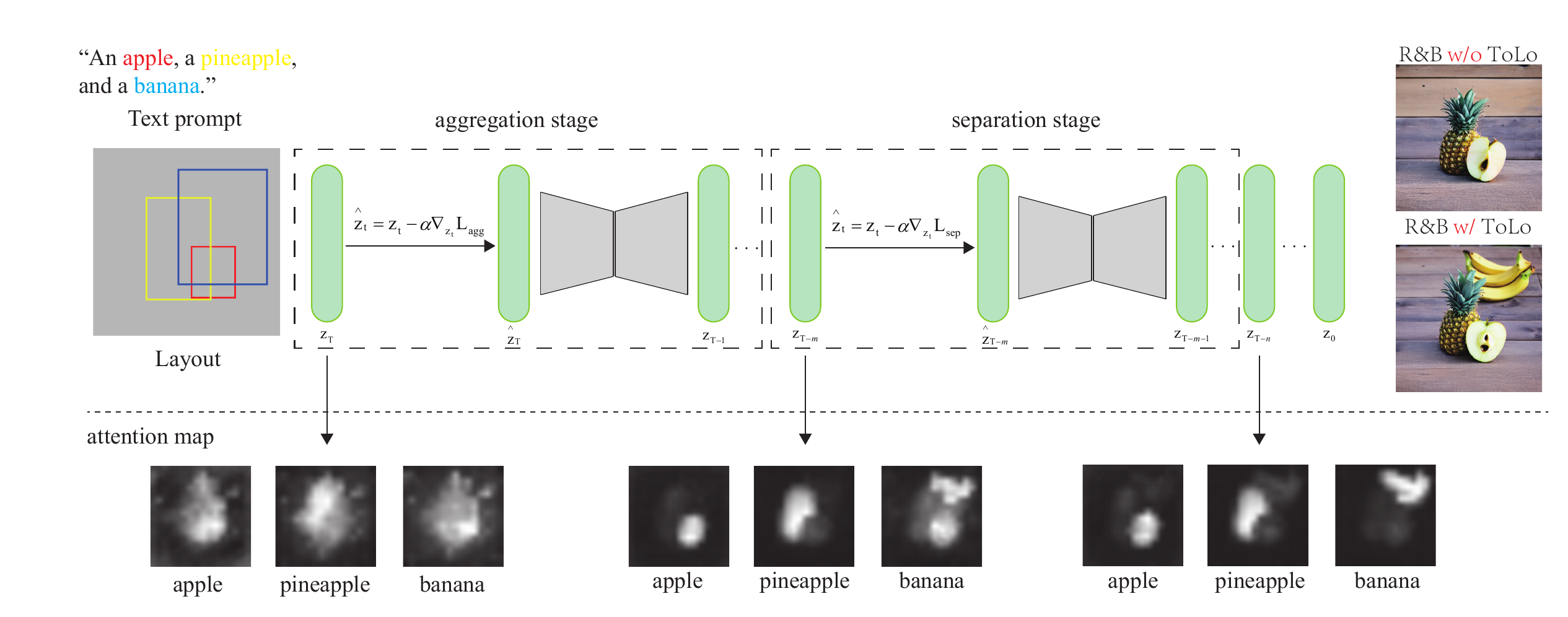}
\end{center}
   \caption{{\bf Overview of ToLo.} In aggregation stage, $L_{\rm agg}$ encourages the attention map of each concept to focus on its respective bounding box, achieving
precise control over the spatial position of concept. In separation stage, $L_{\rm sep}$ ensures the separation between attention maps for different concepts, thereby alleviate the overlap problem of attention maps. }
\label{framework}
\end{figure*}
\subsection{Text-to-Image Diffusion Models}

In recent years, diffusion models have made significant advances in image synthesis tasks \cite{DiffusionModelsBeatGANsonImageSynthesis, DenoisingDiffusionProbabilisticModels, DenoisingDiffusionImplicitModels, Score-BasedGenerativeModelingthroughStochasticDifferentialEquations}. These models generate images by starting with noisy images sampled from a standard Gaussian distribution and iteratively denoising them to produce clean, high-quality images. Among these, text-to-image diffusion models, such as Stable Diffusion \cite{High-ResolutionImageSynthesisWithLatentDiffusionModels}, have achieved state-of-the-art performance in the text-to-image generation task, producing high-quality, high-resolution images.

\subsection{Layout Control in Image Generation}

Despite the remarkable performance of text-to-image diffusion models in generating high-quality images, these models struggle to provide fine-grained control over the spatial structure of the generated content. Specifically, it is difficult for text prompts to guide the spatial relationships between objects during image generation. Prior work \cite{BenchmarkingSpatialRelationshipsinText-to-ImageGeneration} highlights that the ability of current text-to-image models to generate multiple objects and accurately represent their spatial relationships is still limited. As a result, there has been considerable research focusing on enhancing the spatial control of text-to-image diffusion models \cite{Training-FreeLayoutControlWithCross-AttentionGuidance, GLIGEN:Open-SetGroundedText-to-ImageGeneration, GroundedText-to-ImageSynthesiswithAttentionRefocusing,LayeredRenderingDiffusionModelforControllableZero-ShotImageSynthesis,HarnessingtheSpatial-TemporalAttentionofDiffusionModelsforHigh-FidelityText-to-ImageSynthesis,RnB:RegionandBoundaryAwareZero-shotGroundedText-to-imageGeneration,BoxDiff:Text-to-ImageSynthesiswithTraining-FreeBox-ConstrainedDiffusion,AddingConditionalControltoText-to-ImageDiffusionModels,LoCo:LocallyConstrainedTraining-FreeLayout-to-ImageSynthesis}.

Methods like GLIGEN \cite{GLIGEN:Open-SetGroundedText-to-ImageGeneration} and ControlNet \cite{AddingConditionalControltoText-to-ImageDiffusionModels} maintain the high performance of pre-trained models by keeping their weights fixed and training additional components to control the spatial structure of the generated images. Other approaches \cite{Training-FreeLayoutControlWithCross-AttentionGuidance, LayeredRenderingDiffusionModelforControllableZero-ShotImageSynthesis,HarnessingtheSpatial-TemporalAttentionofDiffusionModelsforHigh-FidelityText-to-ImageSynthesis, RnB:RegionandBoundaryAwareZero-shotGroundedText-to-imageGeneration,BoxDiff:Text-to-ImageSynthesiswithTraining-FreeBox-ConstrainedDiffusion,LoCo:LocallyConstrainedTraining-FreeLayout-to-ImageSynthesis} adopt a training-free strategy. Some of these methods leverage the attention map, which serves as a powerful representation of the spatial layout and geometry of generated images \cite{Prompt-to-PromptImageEditingwithCrossAttentionControl}. For example, BoxDiff \cite{BoxDiff:Text-to-ImageSynthesiswithTraining-FreeBox-ConstrainedDiffusion} centers the attention map of a specific token within its defined bounding box using Inner-Box loss and Outer-Box loss. Meanwhile, LoCo\cite{LoCo:LocallyConstrainedTraining-FreeLayout-to-ImageSynthesis} improves layout control by incorporating attention maps from the start-of-text token and the end-of-text token (SoT and EoT), which are observed to capture background and foreground information, respectively. To bridge the gap between consecutive attention maps and discrete layout constraints, the R\&B\cite{RnB:RegionandBoundaryAwareZero-shotGroundedText-to-imageGeneration} extends the foreground region associated with concepts to its minimum bounding rectangle (MBR), aligning it with the ground-truth bounding box. Additionally, it proposes a boundary-aware loss to enhance object discriminability within the corresponding regions.

Despite these advancements, few studies have addressed the issue of overlap attention maps when the input layout has high overlap. In such cases, the lack of separation can lead to issues such as attribute leakage and missing entities, as shown in Figure \ref{attentionmap}, which may degrade the quality of the generated image and damage the alignment between the input prompt and the output.

In contrast to prior work, we propose a two-stage training-free layout-to-image generation framework. Our framework not only centralizes the attention map corresponding to a concept within its bounding box but also ensures the separation of attention maps between different concepts. This dual consideration helps mitigate the issues of attribute leakage and missing entities, as shown in Figure \ref{qualitative}, particularly when dealing with layouts that have high overlap.

\section{Method}

This section describes ToLo and how we apply ToLo to the R\&B.

\subsection{Preliminaries}
{\bf Attention map.} The Stable Diffusion model guides image synthesis by learning the association between text features and image features through cross-attention mechanisms. Given a text prompt $p$, a pre-trained CLIP \cite{LearningTransferableVisualModelsFromNaturalLanguageSupervision} encoder, latent features $\boldsymbol{z}_t$, the model uses CLIP to obtain text embedding $\boldsymbol{e}=f_{\rm CLIP}(p)\in {\mathbbm R}^{N\times d_e}$, $N$, $d_e$ are the text length and text embedding dimension. Then project $\boldsymbol{z}_t$ into a query matrix $\boldsymbol{Q}=f_{\rm Q}(\boldsymbol{z}_t)\in {\mathbbm R}^{(H\times W)\times d}$and $\boldsymbol{e}$ into a key matrix $\boldsymbol{K}=f_{\rm K}(\boldsymbol{e})\in {\mathbbm R}^{N\times d}$. $d$ is the feature projection dimension of query and key. The attention map $\boldsymbol{A}={\rm Softmax}(\frac{\boldsymbol{QK}^T}{\sqrt{d}})\in {\mathbbm R}^{(H\times W)\times N}$.

{\bf Problem setup.} We consider the input layout as $k$ bounding boxes $B=\{\boldsymbol{b}_1,\boldsymbol{b}_2,...,\boldsymbol{b}_k\}$ and $k$ corresponding concepts $S=\{s_1,s_2,...,s_k\}$. $\boldsymbol{b}_i$ is a bounding box provided by the user, and $s_i$ is a concept, which constrained by $\boldsymbol{b}_i$. Note that each concept is a subset of the text prompt $p=\{w_1,w_2,...,w_N\}$.
\subsection{ToLo}

Figure \ref{framework} shows an overview of ToLo. Given a prompt $p$, a set of bounding boxes $B$, and the corresponding set of concepts $S$, the process unfolds in two stages: aggregation and separation. The aggregation stage occupies the first $m$ steps of the entire denoising process, while the separation stage occupies the next $n-m$ steps. In the aggregation stage, we compute $L_{\rm agg}$, encouraging the attention map of each
concept to focus on its respective bounding box. Conversely, in the separation stage, we compute $L_{\rm sep}$, ensuring the separation between attention maps for different concepts. The loss is then used to update the latent representation $\boldsymbol{z}_t$ through a gradient step: $\boldsymbol{z}_t\leftarrow \boldsymbol{z}_t-a\nabla_{\boldsymbol{z}_t}L$ during the intervention steps. Algorithm \ref{ToLo} presents the pseudocode for the ToLo. Note that in different implementations of ToLo, $L_{\rm agg}$ and $L_{\rm sep}$ can have different specific forms.

This framework considers both the centralization of attention maps for individual concepts and the separation of attention maps for different concepts. This approach helps alleviate attribute leakage and missing entities while maintaining precise spatial control, thereby improving the performance of existing methods in handling high-overlap layouts, as illustrated in Figure \ref{qualitative} and Figure \ref{attentionmap}.

\begin{algorithm}
    \caption{Denoising Process of ToLo}
    \label{alg:AOA}
    \renewcommand{\algorithmicrequire}{\textbf{Input:}}
    \renewcommand{\algorithmicensure}{\textbf{Output:}}
    \begin{algorithmic}[1]
        \REQUIRE A text prompt $p$, a trained diffusion model $SD$, bounding boxes $B$ and corresponding concepts $S$, the number of guided steps in the aggregation stage $m$, the total intervention steps $n$, the loss scale $\alpha$, the latent representation $\boldsymbol{z}_T$
        \ENSURE Denoised latent $\boldsymbol{z}_0$

        \FOR{each $t \leftarrow T,T-1,...,1$}
            \IF {$t > T - n$}
                \STATE Obtain Cross Attention Map $\boldsymbol{A}$
                \IF {$t > T - m$}
                    \STATE $Loss \leftarrow L_{\rm agg}(\boldsymbol{A},B,S)$
                \ELSE
                    \STATE $Loss \leftarrow L_{\rm sep}(\boldsymbol{A},B,S)$
                \ENDIF
                \STATE $\boldsymbol{z}_t \leftarrow \boldsymbol{z}_t-\alpha \nabla_{\boldsymbol{z}_t}Loss$
            \ENDIF
            \STATE $\boldsymbol{z}_{t-1} \leftarrow SD(\boldsymbol{z}_t,p,t)$
        \ENDFOR
        
        \RETURN $\boldsymbol{z}_0$
    \end{algorithmic}
\label{ToLo}
\end{algorithm}

\begin{table*}
\caption{Comparison with training-free layout-to-image synthesis methods at different IoU.}
\begin{center}
\resizebox{\linewidth}{!}{
\begin{tabular}{cccccccccc}
\hline
\multirow{2}{*}{Method} & &HRS-Spatial&&&HRS-Color&&&HRS-Size& \\
\cmidrule(r){2-4} \cmidrule(r){5-7} \cmidrule(r){8-10}
&${\rm IoU}=0$&$0<{\rm IoU}\leq 0.1$&${\rm IoU}>0.1$&${\rm IoU}=0$&$0<{\rm IoU}\leq 0.1$&${\rm IoU}>0.1$&${\rm IoU}=0$&$0<{\rm IoU}\leq 0.1$&${\rm IoU}>0.1$\\
\hline
Stable Diffusion&12.87&6.61&7.74&14.20&13.65&11.50&20.87&18.75&4.53\\
Boxdiff&19.16&12.40&18.71&22.59&21.16&18.14&25.22&18.75&6.58\\
Layout-guidance&25.31&14.05&13.55&22.38&13.31&11.95&29.57&21.09&4.94\\
R\&B&41.96&\bf{21.49}&15.48&33.16&24.23&22.57&\bf{43.48}&\bf{28.91}&14.81\\
R\&B+ToLo(Ours)&\bf{42.38}&19.01&\bf{21.94}&\bf{33.58}&\bf{27.30}&\bf{26.11}&42.61&26.56&\bf{16.46}\\
\hline
\end{tabular}
}
\end{center}
\label{hrs_IoU}
\end{table*}

\subsection{Applying ToLo to R\&B}

Since R\&B follows the aforementioned one-stage framework, our aggregation loss is equivalent to the loss proposed in \cite{RnB:RegionandBoundaryAwareZero-shotGroundedText-to-imageGeneration}. Details can be found in the appendix \ref{appendix_aggregation}.

For separation loss, we first get the the foreground mask $\boldsymbol{\hat{M}}_i$ same to \cite{RnB:RegionandBoundaryAwareZero-shotGroundedText-to-imageGeneration}. Specifically, we first compute an aggregated attention map $\boldsymbol{M}_i$ for each $s_i$:
\begin{equation}
\boldsymbol{M}_i=\frac{1}{L}\sum_{l=1}^L\sum_{j=1}^N\mathbbm{1}_{s_i}(w_j)\cdot \boldsymbol{A}_l^j
\end{equation}

Where $\boldsymbol{A}_l^j$ is the attention map of the $j^{th}$ word in the prompt at the $l^{th}$ layer and $\mathbbm{1}_{s_i}(\cdot)$ is an indicator function. Since the cross-attention maps $\boldsymbol{A}_l$ at each layer have different resolutions, they are first upsampled to a unified resolution of 64 × 64, and then averaged to obtain $\boldsymbol{M}_i$. After this, $\boldsymbol{\hat{M}}_i$ is computed by follow:
\begin{equation}
\boldsymbol{M}_i^{\rm norm}=\frac{\boldsymbol{M}_i-{\rm min}_{h,w}(\boldsymbol{M}_i) \cdot \mathbbm{1}}{{\rm max}_{h,w}(\boldsymbol{M}_i)-{\rm min}_{h,w}(\boldsymbol{M}_i)}
\end{equation}
\begin{equation}
\begin{aligned}
    \tau_i=&\lambda\frac{\sum_{h,w}\boldsymbol{M}_i^{\rm norm}\odot \boldsymbol{b}_i}{\sum_{h,w} \boldsymbol{b}_i}+\\
    &(1-\lambda)\frac{\sum_{h,w}\boldsymbol{M}_i^{\rm norm}\odot (\mathbbm{1}-\boldsymbol{b}_i)}{\sum_{h,w} (\mathbbm{1}-\boldsymbol{b}_i)}
\end{aligned}
\end{equation}
\begin{equation}
    (\boldsymbol{\hat{M}}_i)_{h,w}=\left\{
\begin{aligned}
1,(\boldsymbol{M}_i^{\rm norm})_{h,w}\geq \tau_i\\
0,(\boldsymbol{M}_i^{\rm norm})_{h,w}< \tau_i\\
\end{aligned}
\right.
\label{eq4}
\end{equation}

Where $\mathbbm{1}$ is a matrix where every element is 1. $\boldsymbol{\hat{M}}_i$ is the foreground region of the $i^{th}$ concept highlighted by the dynamic threshold $\tau_i$.

Then, our $L_{\rm sep}$ is defined as follow:
\begin{equation}
    \boldsymbol{\hat{M}}_i^{\rm norm}=\boldsymbol{\hat{M}}_i \odot \boldsymbol{M}_i^{\rm norm}
\end{equation}
\begin{equation}
    L_{\rm sep}=\frac{1}{C}\sum_{i\neq j}\frac{\sum_{h,w}\boldsymbol{\hat{M}}_i\odot \boldsymbol{\hat{M}}_j^{\rm norm}}{\sum_{h,w}\boldsymbol{\hat{M}}_j^{\rm norm}}
\end{equation}

Where $C$ is the number of pairs $(i,j)$. For each $(i,j)$, this loss measures the overlap between $s_j$'s attention map in the foreground region and $s_i$'s foreground region. Intuitively, reducing this loss can reduce the overlap between attention maps corresponding to different concepts, thereby improving the performance of the model on high overlapping input layouts.

We then define the total loss as follow:
\begin{equation}
    L=\left\{
\begin{aligned}
&L_{\rm agg},t<m\\
&L_{\rm sep},m\leq t<n\\
\end{aligned}
\right.
\end{equation}

Where $m$ is the number of guided steps in the aggregation stage and $n-m$ is the number of guided steps in the separation stage.

Next, we used the loss $L$ to update the latent representation $\boldsymbol{z}_t$ with a gradient step $\boldsymbol{z}_t\leftarrow \boldsymbol{z}_t-a\nabla_{\boldsymbol{z}_t}L$.

\section{Experiments}
\subsection{Experimental Setup}
{\bf Datasets.}  We use HRS-Bench \cite{HRS-Bench:Holistic}, a comprehensive benchmark for text-to-image models, which provides a variety of prompts categorized into three main topics: accuracy, robustness, and generalization. Since our method focuses on layout control, we specifically select three categories from HRS-Bench: spatial relationships, size, and color. The number of prompts for each category is 1002, 501, and 501, respectively. To evaluate the performance of our method with varying degrees of overlap in layout, we divide HRS-Bench into three parts based on the IoU. Specifically, for each input layout, we calculate the maximum IoU as its representative value. Formally, we take ${\rm max}_{\boldsymbol{b}_i,\boldsymbol{b}_j\in B,i\neq j}({\rm IoU}(\boldsymbol{b}_i,\boldsymbol{b}_j))$ as the IoU. Based on this, we categorize HRS-Bench into three groups: ${\rm IoU}=0$, $0<{\rm IoU}<=0.1$, ${\rm IoU}>0.1$. Since HRS-Bench does not provide layout information, we incorporate publicly available layout labels from Phung et al. \cite{GroundedText-to-ImageSynthesiswithAttentionRefocusing} for partitioning and evaluation. The number of prompts in each category of the dataset is provided in Appendix \ref{appendix_data}.

{\bf Evaluation Metrics.} We follow the standard evaluation protocol of HRS. Specifically, we employ the pre-trained UniDet \cite{SimpleMulti-DatasetDetection}, a multi-dataset detector, on all synthesized images. For spatial relationships, the predicted bounding boxes are employed to assess whether the spatial relations are correctly grounded. For size composition, the sizes of the predicted bounding boxes are used to verify the correctness of the size composition. For color composition, we first convert the image to the hue color space, then calculate the average hue value within each bounding box and compare it to the predefined color space.

{\bf Implementation Details.} We use the official Stable Diffusion v1.4 \cite{High-ResolutionImageSynthesisWithLatentDiffusionModels}, which was trained on the LAION-5B \cite{LAION-5B:Anopenlarge-scaledatasetfortrainingnextgenerationimage-textmodels} dataset, as the base text-to-image synthesis model. For the hyperparameters, we set $m=10$, $\alpha=70$, and $n=12$. Note that $\alpha$, along with all other hyperparameters, is the same as specified in the code published by  \cite{RnB:RegionandBoundaryAwareZero-shotGroundedText-to-imageGeneration}.

\begin{table}
\caption{Comparison with training-free layout-to-image synthesis methods on HRS-Bench. This result is obtained by taking a weighted average of the results in Table \ref{hrs_IoU}.}
\begin{center}
\resizebox{\linewidth}{!}{
\begin{tabular}{cccc}
\hline
Method&HRS-Spatial&HRS-Color&HRS-Size\\
\hline
Stable Diffusion&11.30&13.74&12.14\\
Boxdiff&18.26&21.74&14.20\\
Layout-guidance&22.10&19.38&15.02\\
R\&B&35.32&30.16&\bf{25.31}\\
R\&B+ToLo(Ours)&\bf{36.33}&\bf{31.47}&\bf{25.31}\\
\hline
\end{tabular}
}
\end{center}
\label{hrs}
\end{table}

\begin{figure*}
\begin{center}
\includegraphics[width=1\linewidth]{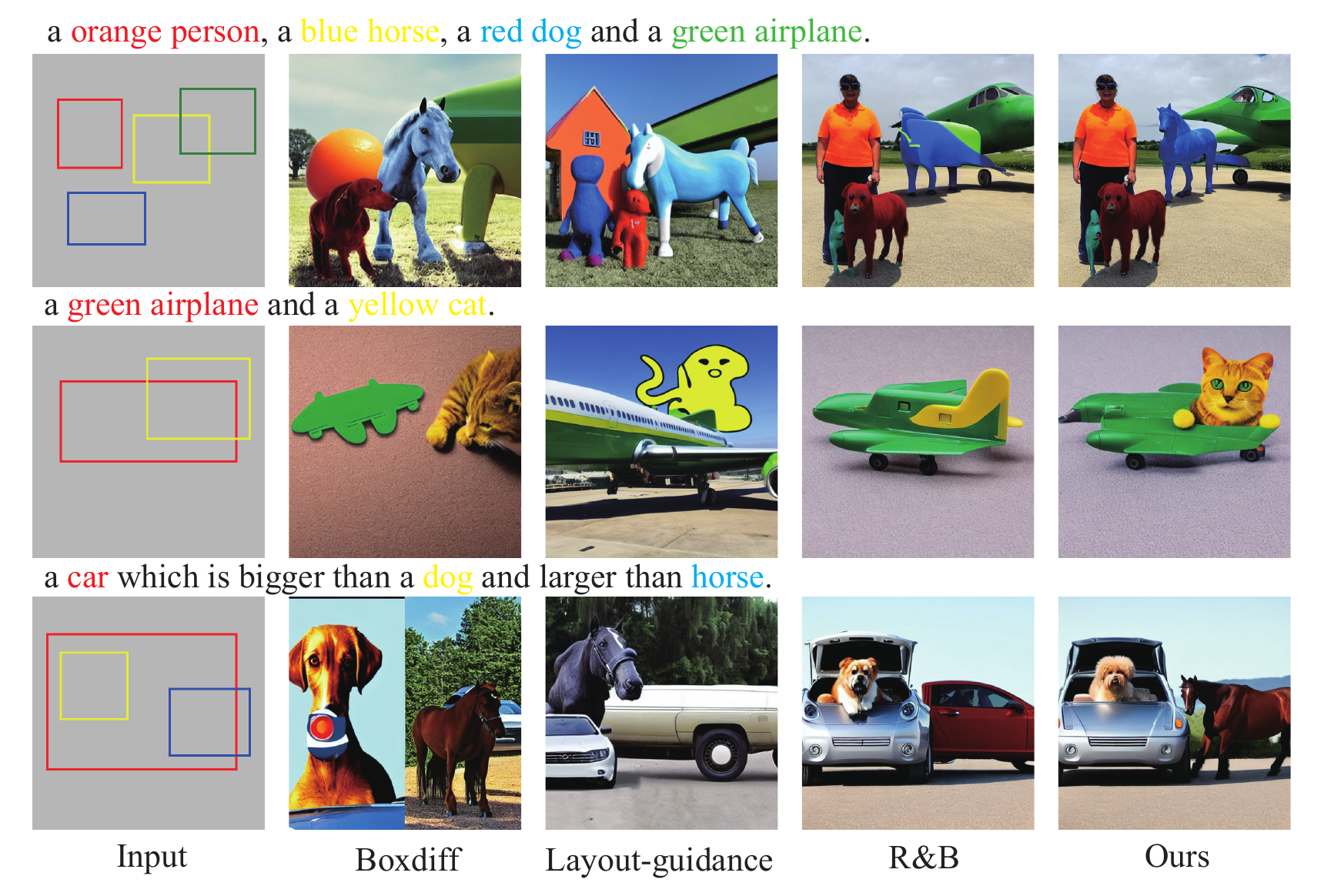}
\end{center}
   \caption{{\bf Qualitative result.} All methods take the same grounded texts as inputs. The results show that our proposed ToLo can effectively alleviate problems such as attribute leakage and missing entities.}
\label{qualitative2}
\end{figure*}

\subsection{Quantitative Results}
We compare our method with various state-of-the-art layout-to-image approaches based on the Stable Diffusion v1.4, as shown in Tables \ref{hrs_IoU} and \ref{hrs}. For the high IoU, our method shows a remarkable performance boost compared to the existing methods. In the spatial category, our method outperforms the R\&B by 6.46\% on the ${\rm IoU}>0.1$ subset of HRS. This is because we alleviate the issue of attribute leakage and missing entities in existing methods, as shown in Figure \ref{qualitative2}, by separating the attention maps. However, we observed that the separation stage may impair the ability to control the size of the object. As shown in Figure \ref{attentionmap}, although our method successfully generates a yellow clock, the resulting object is smaller than the given bounding box due to the separation between the yellow clock and the red apple. This may result in our method performing less effectively when the input layout has low overlap in the size category. However this limitation can be easily addressed by introducing the IoU threshold. Specifically, for ${\rm IoU}>k$, the two-stage method is used, while for ${\rm IoU}\leq k$, the original one-stage method is employed, with $k$ representing the IoU threshold.
\label{Quantitative Results}
\subsection{Qualitative Result}
Figure \ref{qualitative2} presents a visual comparison of various state-of-the-art methods, highlighting that our proposed ToLo effectively mitigates issues such as attribute leakage and missing entities, thereby improving the performance of existing methods in handling high-overlap layouts.

For instance, in the first row, Boxdiff and Layout-guidance fail to generate the green airplane and the orange person, suffering from object disappearance, while R\&B fails to generate the blue horse due to attribute leakage. In contrast, our method successfully generates all the objects. In the third row, Boxdiff fails to generate a car, and Layout-guidance fails to generate a dog, both encountering the issue of object disappearance. R\&B generates a car with the color of a horse, exhibiting attribute leakage. In contrast, our method accurately generates a horse and a dog.
\label{Qualitative Result}

\begin{table}
\caption{Compare with one-stage method at different loss scales.  In this experiment, we fixed the total number of intervention steps to 12. For our method, we set $m=10$ and $n=12$.}
\begin{center}
\resizebox{\linewidth}{!}{
\begin{tabular}{cccccc}
\hline
\multirow{2}{*}{Method}&\multicolumn{5}{c}{Loss Scale}\\
\cmidrule(r){2-6}
&50&60&70&80&90\\
\hline
One Stage&20.00&21.29&18.06&18.06&20.00\\
R\&B+ToLo(Ours)&\bf{23.23}&\bf{23.87}&\bf{21.94}&\bf{20.65}&\bf{23.23}\\
\hline
\end{tabular}
}
\end{center}
\label{ablation_scale}
\end{table}
\subsection{Ablation Studies}

To further illustrate the robustness and effectiveness of ToLo, we replace the separation stage in our method with the aggregation stage to obtain a corresponding one-stage method. In this configuration, the one-stage method has the same number of intervention steps as our method. In table \ref{ablation_scale}, we compare the performance of our method with the corresponding one-stage method on the ${\rm IoU} > 0.1$ subset of the HRS-spatial dataset, across different loss scales. The results demonstrate that our method is highly robust and consistently outperforms the one-stage method across various loss scales.
\section{Conclusion}
In this paper, we identify a critical issue in existing methods: they fail to adequately separate the attention maps of different concepts. This limitation becomes particularly problematic when input layouts contain significant overlap, as it causes the attention maps of distinct concepts to merge, leading to attribute leakage and the omission of entities.

To address these challenges, we introduce ToLo, a two-stage training-free layout-to-image generation framework, and apply it to the R\&B. To provide a more effective evaluation, we divide HRS-Bench into three parts based on the IoU, creating a new dataset for layout-to-image
generation with varying levels of overlap. Experimental results demonstrate that our proposed framework significantly enhances
the performance of existing baselines, both quantitatively and qualitatively, on high-overlap layouts, while also exhibiting strong robustness.

{\small
\bibliographystyle{ieee}
\bibliography{main}
}

\clearpage

\appendix
\section{More examples of overlapping attention maps}
\label{appendix_attentionmap}

\begin{figure}[H]
\begin{center}
   \includegraphics[width=1\linewidth]{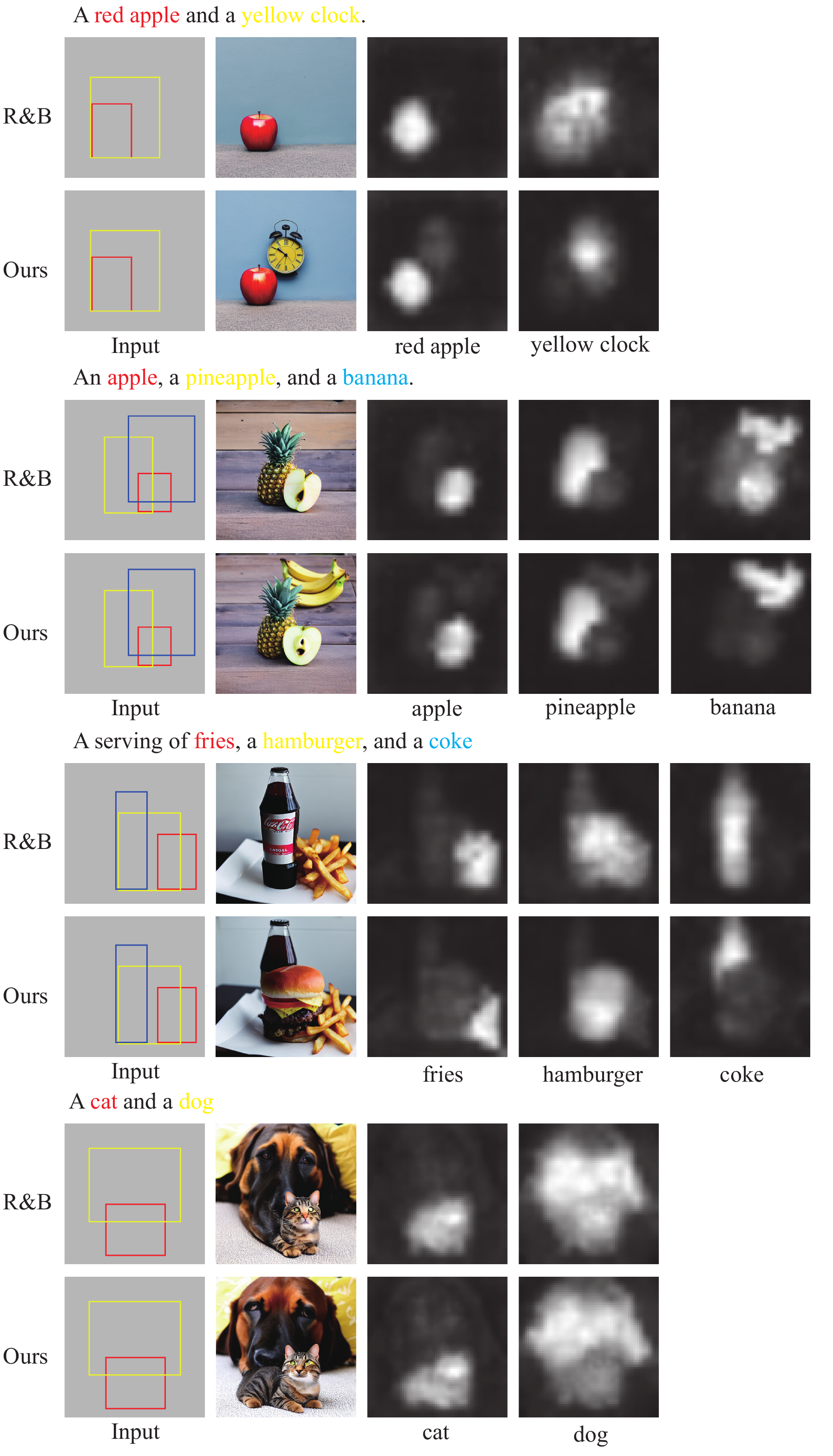}
\end{center}
\caption{More examples of overlapping attention maps.}
\label{appendix}
\end{figure}

\section{Details of the dataset}
\label{appendix_data}

The number of prompts in each part of our dataset is shown in Table \ref{hrs_spatial_number} to \ref{hrs_size_number}. Note that we have removed some dirty data. Specifically, we excluded any layout where the bounding box $\boldsymbol{b}_i$ is represented by $\{x_{\rm min}, y_{\rm min}, x_{\rm max}, y_{\rm max}\}$, if any of the following conditions were met:
\begin{itemize}
    \item $\max(x_{\rm min}, y_{\rm min}, x_{\rm max}, y_{\rm max}) > 512$
    \item $x_{\rm min} \geq x_{\rm max}$
    \item $y_{\rm min} \geq y_{\rm max}$
\end{itemize}

\begin{table} [H]
\begin{center}
\resizebox{\linewidth}{!}{
\begin{tabular}{cccc}
\hline
\multirow{2}{*}{} & &HRS-Spatial& \\
\cmidrule(r){2-4}
&${\rm IoU}=0$&$0<{\rm IoU}\leq 0.1$&${\rm IoU}>0.1$\\
\hline
Number&715&121&155\\
\hline
\end{tabular}
}
\end{center}
\caption{The number of prompts in spatial category.}
\label{hrs_spatial_number}
\end{table}

\begin{table} [H]
\begin{center}
\resizebox{\linewidth}{!}{
\begin{tabular}{cccc}
\hline
\multirow{2}{*}{} & &HRS-Color& \\
\cmidrule(r){2-4}
&${\rm IoU}=0$&$0<{\rm IoU}\leq 0.1$&${\rm IoU}>0.1$\\
\hline
Number&342&87&67\\
\hline
\end{tabular}
}
\end{center}
\caption{The number of prompts in color category.}
\label{hrs_color_number}
\end{table}

\begin{table} [H]
\begin{center}
\resizebox{\linewidth}{!}{
\begin{tabular}{cccc}
\hline
\multirow{2}{*}{} & &HRS-Size& \\
\cmidrule(r){2-4}
&${\rm IoU}=0$&$0<{\rm IoU}\leq 0.1$&${\rm IoU}>0.1$\\
\hline
Number&115&128&243\\
\hline
\end{tabular}
}
\end{center}
\caption{The number of prompts in size category.}
\label{hrs_size_number}
\end{table}

\section{Aggregation loss}

{\bf Box selection.} To bridge the domain gap between $\boldsymbol{\hat{M}}_i$ in Eq.(\ref{eq4}) and the ground-truth bounding box $\boldsymbol{b}_i$, we perform box selection by extending $\boldsymbol{\hat{M}}_i$ to its minimum bounding rectangle (MBR) $\boldsymbol{\hat{b}}_i$ to match $\boldsymbol{b}_i$. However, directly optimizing $\boldsymbol{\hat{b}}_i$ towards its ground-truth $\boldsymbol{b}_i$ is not feasible because the MBRs are hard and non-differentiable masks. Towards practical implementation, we use a variant of $\boldsymbol{\hat{b}}_i$:
\begin{equation}
    \boldsymbol{\hat{b}}_i^a={\rm stopgrad}(\boldsymbol{\hat{b}}_i-\boldsymbol{M}_i^{\rm norm})+\boldsymbol{M}_i^{\rm norm}
\end{equation}

{\bf Region-aware loss for attention guidance.} The region-aware loss is defined as follow:
\begin{equation}
\boldsymbol{M}_i^s={\rm sigmoid}(s\cdot (\boldsymbol{M}_i^{\rm norm} - \tau_i\cdot \mathbbm{1}))
\end{equation}
\begin{equation}
    IoU_i=\frac{\sum_{h,w}\boldsymbol{\hat{b}}_i\odot \boldsymbol{b}_i}{\sum_{h,w}(\boldsymbol{\hat{b}}_i+(\mathbbm{1}-\boldsymbol{\hat{b}}_i)\odot \boldsymbol{b}_i)}
\end{equation}
\begin{equation}
\begin{aligned}
    L_{\rm r}^i=&(1-IoU_i)\cdot (\lambda_s(1-\frac{\sum_{h,w}\boldsymbol{M}_i^s\odot \boldsymbol{b}_i}{\sum_{h,w}\boldsymbol{M}_i^s})+\\
    &\lambda_a(1-\frac{\sum_{h,w}\boldsymbol{\hat{b}}_i^a\odot \boldsymbol{b}_i}{\sum_{h,w}\boldsymbol{\hat{b}}_i^a}))
\end{aligned}
\end{equation}

{\bf Boundary-aware loss for attention guidance.} The boundary-aware loss is defined as follow:
\begin{equation}
    \boldsymbol{\varepsilon}_i={\rm Sobel}(\boldsymbol{M}_i)
\end{equation}
\begin{equation}
    L_{\rm b}^i=(1-IoU_i)\cdot (1-\frac{\sum_{h,w}\boldsymbol{\varepsilon}_i\odot \boldsymbol{b}_i}{\sum_{h,w}\boldsymbol{\varepsilon}_i})
\end{equation}

Where $Sobel(\cdot)$ represents the Sobel Operator and $\boldsymbol{\varepsilon}_i$ is the extracted edge map from the aggregated attention map.

{\bf Aggregation loss.} Our $L_{\rm agg}$ is defined as follow:
\begin{equation}
    L_{\rm agg}=\sum_{\boldsymbol{b}_i\in B} L_{\rm r}^i+L_{\rm b}^i
\end{equation}
\label{appendix_aggregation}
\end{document}